\documentclass[10pt,twocolumn,letterpaper]{article}

\usepackage{cvpr}              

\usepackage{graphicx}
\usepackage{amsmath}
\usepackage{amssymb}
\usepackage{booktabs}
\usepackage{multirow}
\usepackage{makecell}
\usepackage{bbding}

\usepackage[pagebackref,breaklinks,colorlinks]{hyperref}

\usepackage[capitalize]{cleveref}
\crefname{section}{Sec.}{Secs.}
\Crefname{section}{Section}{Sections}
\Crefname{table}{Table}{Tables}
\crefname{table}{Tab.}{Tabs.}

\def\cvprPaperID{5638} 
\def\confName{CVPR}
\def\confYear{2022}

\begin{document}

\title{Towards Accurate Facial Landmark Detection via Cascaded Transformers}

\author{
	Hui Li\thanks{The first two authors equally contributed to this work. H.Li is the corresponding author.} \textsuperscript{\rm 1}, ~~
	Zidong Guo$^{\ast}$\textsuperscript{\rm 1}, ~~
	Seon-Min Rhee\textsuperscript{\rm 2},  ~~
	Seungju Han\textsuperscript{\rm 2},  ~~
    Jae-Joon Han\textsuperscript{\rm 2}\\
	\textsuperscript{\rm 1}Samsung R\&D Institute China Xi'an (SRCX) \\
	\textsuperscript{\rm 2}Samsung Advanced Institute of Technology (SAIT), South Korea \\
	{\tt\small {hui01.li, zidong.guo, s.rhee, sj75.han, jae-joon.han}@samsung.com}
}

\maketitle

\begin{abstract}
Accurate facial landmarks are essential prerequisites for many tasks related to human faces. In this paper, an accurate facial landmark detector is proposed based on cascaded transformers. We formulate facial landmark detection as a coordinate regression task such that the model can be trained end-to-end. With self-attention in transformers, our model can inherently exploit the structured relationships between landmarks, which would benefit landmark detection under challenging conditions such as large pose and occlusion. During cascaded refinement, our model is able to extract the most relevant image features around the target landmark for coordinate prediction, based on deformable attention mechanism, thus bringing more accurate alignment. In addition, we propose a novel decoder that refines image features and landmark positions simultaneously. With few parameter increasing, the detection performance improves further. Our model achieves new state-of-the-art performance on several standard facial landmark detection benchmarks, and shows good generalization ability in cross-dataset evaluation.

\end{abstract}

\section{Introduction}
\label{sec:intro}

Facial landmark detection aims to automatically localize fiducial facial landmark points on human faces. It serves as an essential step for several facial analysis tasks, such as face recognition, facial expression analysis, face frontalization and 3D face reconstruction~\cite{survey1}.

Facial landmark detection has received significant improvement in recent years. Existing approaches mainly fall into two categories, i.e., coordinate regression-based methods and heatmap-based methods. Coordinate regression-based methods~\cite{Wing,ODN} map the input image to landmark coordinates via fully connected prediction layers.
To improve accuracy, coordinate regression is usually cascaded as a coarse-to-fine manner~\cite{decafa, DAG2020} or integrated with heatmap regression module~\cite{LAB,HGs}.
Heatmap-based methods~\cite{AWing,HRNet2019,HSLE}
usually predict heatmaps by fully convolutional networks and then obtain the landmarks according to the peak probability locations on the heatmaps. Since heatmap-based models can preserve the spatial structure of image features, they have better performance than coordinate regression-based models generally.

\begin{figure}[t]
    \centering
        \includegraphics[width=0.9\linewidth]{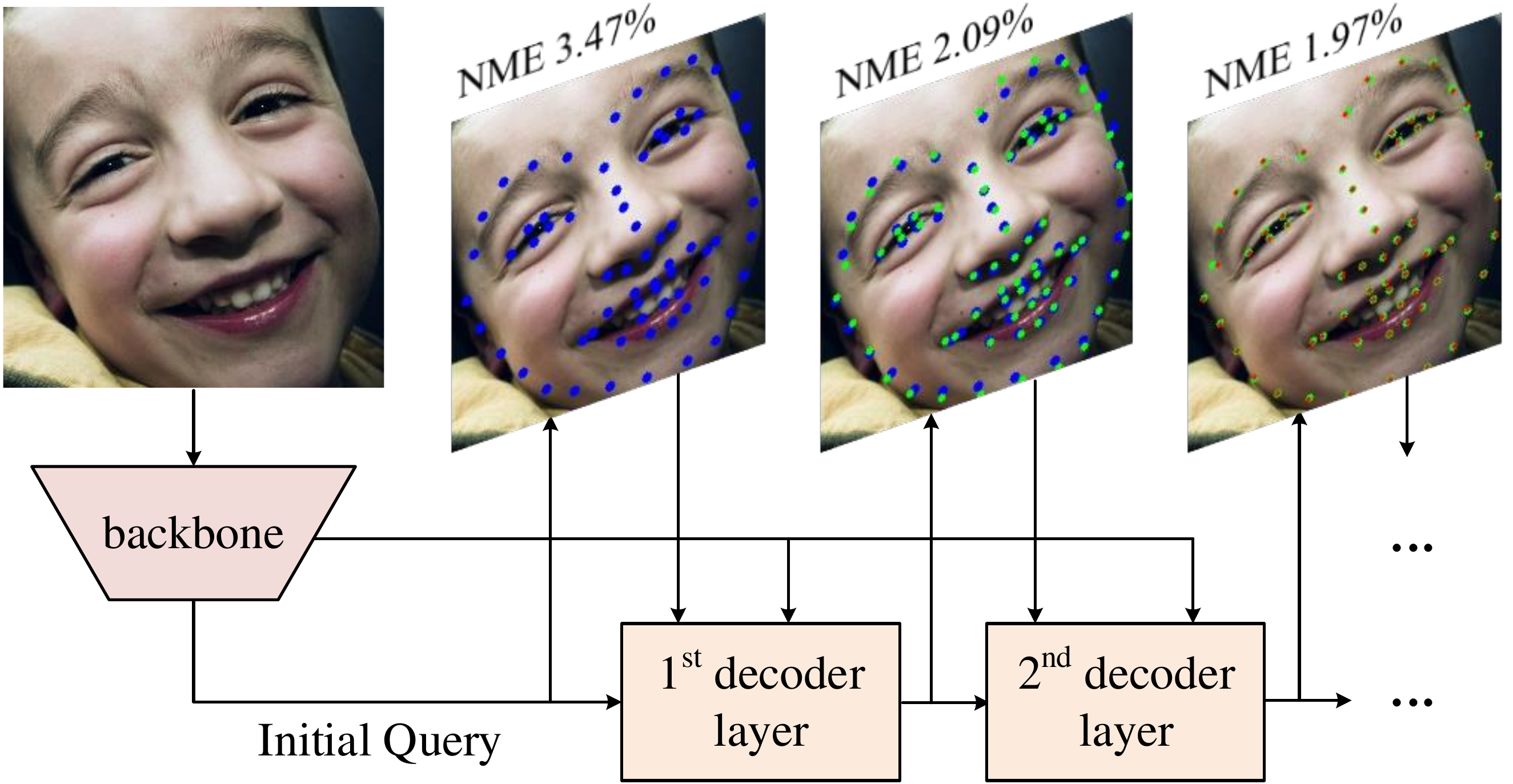}
    \caption{Illustration of our entire framework. The initial query and landmark locations are generated based on the image features, and are continuously updated along with the decoding process.}
\label{framework0}
\vspace{-4mm}
\end{figure}

Although heatmap-based methods have relatively higher detection accuracy, they suffer from three major issues. 1) The required post-processing step is non-differentiable, which disables the end-to-end training.
2) Considering the computational complexity, the resolution of heatmaps is usually lower than that of the input images, resulting in a quantization error inevitably and limiting the performance.
3) They concern more on local texture information and neglect global sensing on face shape, making them vulnerable to large appearance variation such as occlusions.

In contrast, coordinate regression based methods can bypass the aforementioned drawbacks and enable end-to-end model training.
However, the fully connected layers destroy the spatial structure of local image features, which deteriorates the localization performance greatly~\cite{PIPNet2021}.

In this work, we propose a coordinate regression-based model, Deformable Transformer Landmark Detector (DTLD), for accurate facial landmark detection. On one hand, our model avoids the aforementioned shortcomings of heatmap-based methods, and can be well-trained end-to-end, without heuristical post-processing. On the other hand, the model is capable of extracting the most relevant features from multi-level feature maps around the target landmark for coordinate prediction, which preserves the local spatial structure and improves the localization accuracy to a large extent. Moreover, our method helps to exploit the underlying relationship among landmarks and incorporate rich structure knowledge, which enables a robust model to tackle various scenarios such as expression or occlusion.

Inspired by the great success of DEtection TRansformer (DETR) in object detection and keypoint detection~\cite{DETR2020,deformable2020, TFPose}, we formulate the landmark detection as a gradually refined N-coordinate prediction task, where N is the number of facial landmarks. Self-attention block is adopted to learn potential structural dependencies.
Then multi-scale image feature based deformable attention~\cite{deformable2020} is employed, where landmark related information is used as the guidance to adaptively extract the most relevant features and refine the coordinates.
Different from~\cite{DETR2020,deformable2020} that define redundant object queries and use bipartite matching to classify objects, here the number of queries is set to be the number of landmarks exactly, following the practice in~\cite{TFPose, BarrelNet}, which simplifies the training process largely. Instead of using randomly initialized query embedding and similar to the DQInit proposed in~\cite{BarrelNet}, we design a more meaningful image-related query-initialization method, which provides coarse landmark locations rather than a fiducial landmark template. Different from~\cite{TFPose,BarrelNet}, we further explore a parallel decoder where both image features and landmark coordinates are refined simultaneously in the decoding process. It improves the detection performance further. The entire framework is illustrated in Figure~\ref{framework0}.

The main contributions can be summarized as follows.

1) We propose a coordinate regression-based facial landmark detector DTLD by cascaded deformable transformers, based on Deformable DETR~\cite{deformable2020}. DTLD could iteratively capture structural relationships among landmarks and the most relevant visual contextual information to achieve efficient and effective detection.

2) A parallel decoder is further explored to enhance the detection accuracy, with few model parameter increasing.

3) We conduct extensive experiments to analyze the effectiveness of the proposed method, by both quantitative evaluations and qualitative visualizations. Our model contributes to tackle landmark detection under various scenarios. 
It achieves new state-of-the-art (SOTA) accuracy on several facial landmark detection benchmarks, and shows good generalization ability in cross-dataset evaluation.

\section{Related work}
\subsection{Facial Landmark Detection}
As stated above, the existing approaches on facial landmark detection can be roughly divided into two categories. 

\noindent{\bf Heatmap-based} methods usually use high-resolution feature maps for precise localization and achieve encouraging performance. Stacked hourglass network~\cite{hourglass2016} and U-Net~\cite{unet2015} are two typical architectures that perform well in heatmap-based methods~\cite{HSLE,ADA, AWing, hourglass2, unet1}.
Specifically, HSLE~\cite{HSLE} proposes to hierarchically depict holistic and local structures obtained by stacked hourglass network for accurate alignment.
LUVLi~\cite{LUVLi} investigates U-Net for jointly predicting landmark locations, associated uncertainties of these predicted locations and landmark visibilities. HRNet~\cite{HRNet2019} also shows promising results by connecting and exchanging information via fusing multi-scale image features across multiple branches to obtain high-resolution maps. More recently, PIPNet~\cite{PIPNet2021} conducts heatmap and offset predictions simultaneously on low-resolution feature maps, which largely reduces inference time and achieves competitive accuracy.

\noindent{\bf Coordinate regression-based} models are mostly fast, but not accurate enough~\cite{PIPNet2021}. In order to improve the accuracy, most algorithms are designed to make predictions in a coarse-to-fine manner through a cascaded structure~\cite{decafa,stackedcoord2,stackedcoord3,Wing}. For instance, Dapogny~\etal~\cite{decafa} proposed DeCaFA that uses fully convolutional U-net to preserve the full spatial resolution throughout the cascaded regression for accurate face alignment.
LAB~\cite{LAB} was proposed by predicting facial boundary as a geometric constraint via heatmap regression to help landmark coordinate prediction. Li~\etal~\cite{DAG2020} adopt a cascaded Graph Convolutional Network to dynamically leverage global and local features for precise prediction. Although this method shows superior performance, it relies more on high-resolution feature maps which is computationally expensive. A similar work proposed recently is BarrelNet~\cite{BarrelNet} which adapts DETR to landmark detection and set the number of query to be exactly fixed as the number of landmark points. It also proposes to use dynamic query from the input image features (DQInit) for better performance. The difference between the two work are as follows: 1) Different from our cascaded refinement process, BarrelNet predicts landmarks after the last decoder layer directly. 2) Instead of exploring multi-level image features, BarrelNet adopts the last backbone feature and proposes a QAMem module to improve the accuracy. 3) BarrelNet calculates the dynamic query after global average pooling on memory feature, whereas our design extracts initial query from the spatial dimension of image feature, and calculates coarse initial landmark coordinates further to constrain them to be landmark related. 
4) BarrelNet shows that more encoders harm the detection accuracy. Nevertheless, our experiments show that both more encoder and more decoder layers contribute to higher detection performance. Hence, we further propose a parallel decoder to keep the effect of encoder while saving parameters.

\subsection{Transformers in Vision Tasks}


Attention mechanism in transformer~\cite{transformer2017} is able to encode distant dependencies or heterogeneous interactions, and has shown outstanding performance on lots of computer vision tasks~\cite{Vit2020,PVT,DETR2020,deformable2020}. VIT~\cite{Vit2020} is the first that employs pure transformer for image classification. PVT~\cite{PVT} integrates pyramid feature maps and spatial property into the model design. 
DETR~\cite{DETR2020} and Deformable DETR~\cite{deformable2020} view object detection as a direct set prediction task and formulate object detection to be trained end-to-end.
Yang~\etal~\cite{TransPos} introduced transformer for human pose estimation, and employed attention layers to capture long-range spatial dependencies between human body parts. The model is still heatmap-based. Li~\etal~\cite{PoseTrans} proposed pose recognition transformer based on DETR. However, it still needs to perform keypoint detection by finding a match between numerous predictions and the ground-truth. TFPose~\cite{TFPose} proposed to regress keypoint coordinates directly based on deformable DETR. It still inherits the encoder-decoder architecture, while we propose a parallel decoder in addition. With a small amount of parameters and computation, our model achieves the highest accuracy on facial landmark detection.


\begin{figure}[t]
    \centering
        \includegraphics[width=0.9\linewidth]{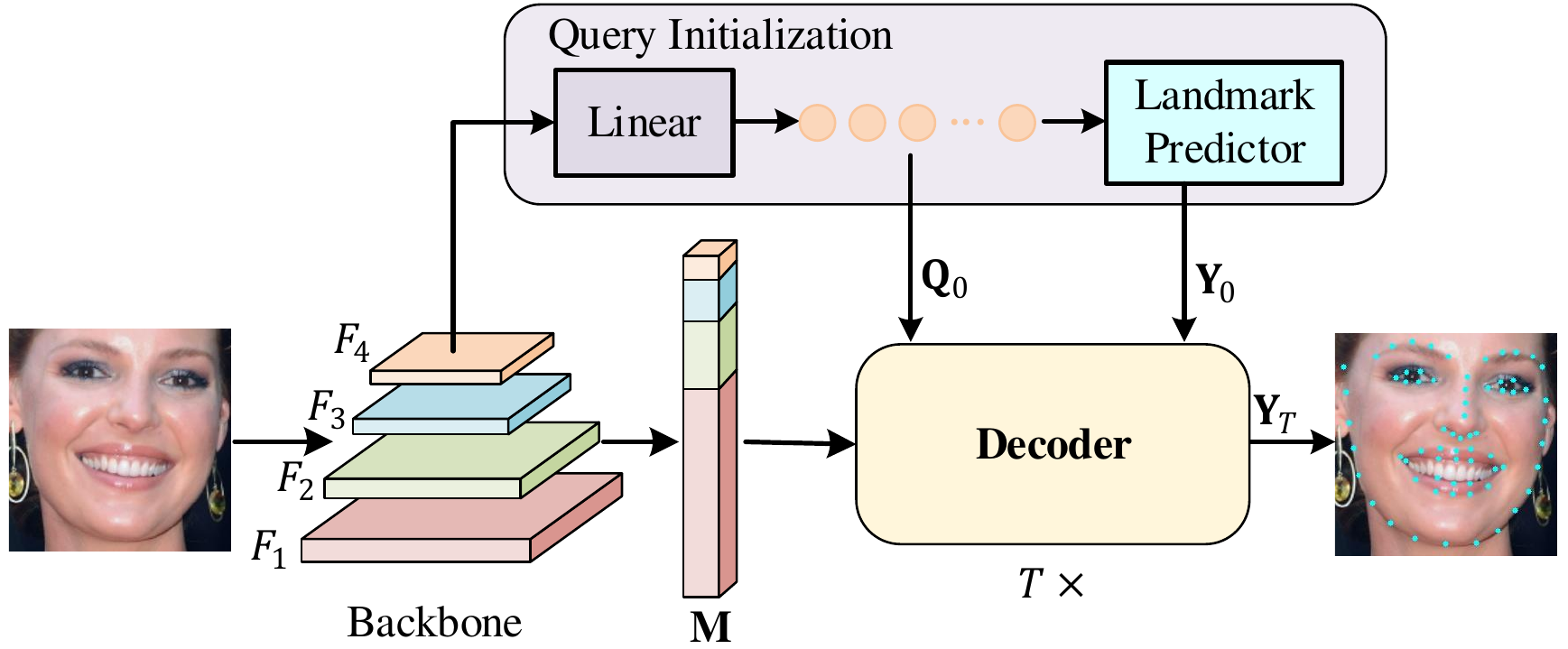}
    \caption{The architecture of our proposed \textbf{DTLD}. $\textbf{Q}_0$ is obtained from $\textbf{F}_4$, the last layer of backbone features, through a linear projection on spatial dimension, and is further transformed into initial landmark coordinates $\textbf{Y}_0$, which are adjusted by $T$ decoder layers to get the final positions $\textbf{Y}_T$.}
\label{framework}
\vspace{-2mm}
\end{figure}

\section{Method}
\label{sec:model}

The architecture of the proposed DTLD
is presented in Figure~\ref{framework}. It is composed of a backbone network for image feature extraction, a query initialization module, and a decoder module for landmark prediction. We adopt a cascaded regression framework where the coordinate offsets are predicted by each decoder layer. The landmark coordinates are refined iteratively during the decoding process. We introduce each part in detail in the following.


\subsection{Backbone}
The backbone contains an ImageNet~\cite{AlexNet2012} pre-trained ResNet-18~\cite{ResNet2016}. Pyramid features are output, which are denoted as $F_1, F_2, F_3, F_4$, with down-sampling ratios of ${4,8,16,32}$ relative to the input image. A $1\times1$ convolution is followed to project the features into the same number of channels.
These features are then flattened and concatenated together, and will be used as the memory feature for decoder, denoted as $\textbf{M} \in \mathcal{R}^{M \times C}$, where $M$ is the length of the flattened features.

\subsection{Query Initialization}

A learnable query matrix $\textbf{Q}$ is defined in~\cite{DETR2020,deformable2020}, which is randomly initialized and updated to represent object related information. In our model, the query matrix $\textbf{Q}$ is defined to have the size of $N \times C$, where $N$ is the number of landmarks and $C$ is the feature dimension.
Rather than random initialization, we extract $N$ features from $F_4$ by a linear projection on spatial dimension, and use them as the initial query features, \ie.
\vspace{-1mm}
\begin{equation}
\label{q0}
\textbf{Q}_0=FC(F_4 ^\mathsf{T})^\mathsf{T},
\vspace{-1mm}
\end{equation}
We reuse $F_4$ to denote the flattened feature, $\textbf{Q}_0 \in \mathcal{R}^{N \times C}$.

The obtained initial query features are expected to be landmark-related. A landmark predictor (another linear projection layer followed by Sigmoid in this paper) is employed to transform them into $N$ landmark coordinates, \ie, 
\begin{equation}
\vspace{-1mm}
\label{y0}
\textbf{Y}_0= \sigma (FC(\textbf{Q}_0)),
\end{equation}
where $\textbf{Y}_0 \in \mathcal{R}^{N \times 2}$ are the initial landmark coordinates, which will be used as the initial reference points for feature sampling in decoding process as well.

\begin{figure}[t]
    \centering
        \includegraphics[width=0.9\linewidth]{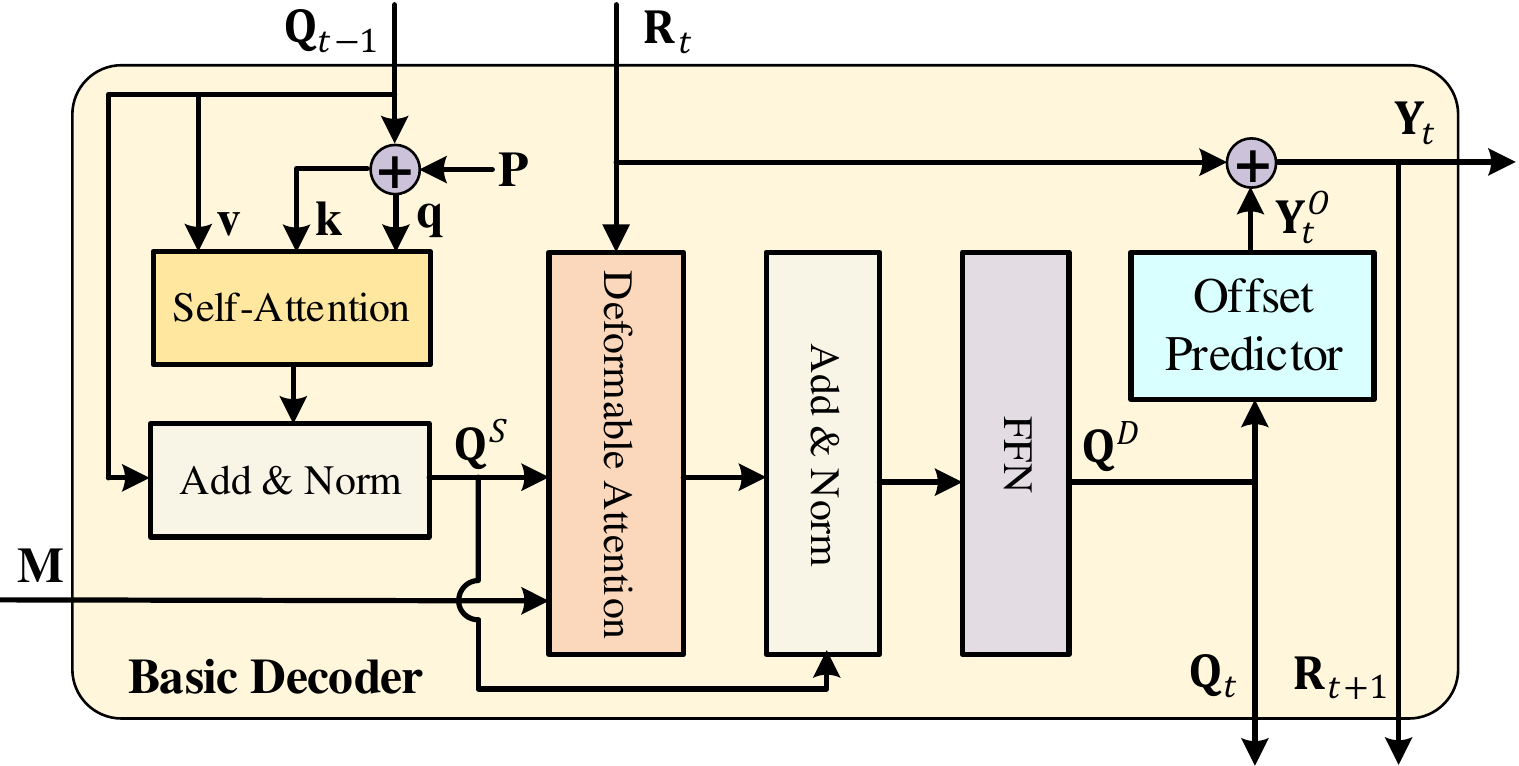}
    \caption{Detailed illustration of the basic decoder.  Memory feature \textbf{M} and the previous \textbf{Q} jointly participate in updating the landmark positions on the basis of previous coordinates \textbf{R}. For the first decoder layer, $\textbf{R}_1$ is the initial landmark location $\textbf{Y}_0$.}
\label{basicdecoder}
\vspace{-2mm}
\end{figure}

\subsection{Decoder Module}
The decoder module is composed of $T$ decoder layers. Each layer takes as inputs a query matrix $\textbf{Q}$, a memory feature $\textbf{M}$ and reference points $\textbf{R}$, and outputs landmark coordinate offsets in regards to $\textbf{R}$. Based on whether updating the memory feature, two types of decoders are explored, \ie, a basic one and a parallel one. They work independently. The former is simple and efficient, setting a strong baseline for landmark detection, while the latter presents a slightly higher detection accuracy.

\noindent{\bf Basic Decoder.}
The configuration of the basic decoder layer is illustrated in Figure~\ref{basicdecoder}.
It mainly consists of a self-attention layer, a deformable attention layer, and an offset predictor.

Specifically, the self-attention layer only adopts the query matrix $\textbf{Q}$ as input. It learns the structure dependency among landmarks by dense interactions.
This information is image-independent intrinsically, where facial attributes like pose and expression will be captured and these attributes have been proven to be important for landmark localization~\cite{DAG2020,TFPose}.
The self-attention layer takes $\textbf{Q}^P$,$\textbf{Q}^P$,\textbf{Q} as query, key and value separately, and $\textbf{Q}^P=\textbf{Q}+\textbf{P}$, where \textbf{P} is a learnable position embedding.
The output from self-attention layer is denoted as $\textbf{Q}^S = [\textbf{q}_1^S, \dots, \textbf{q}_N^S]$, where
\vspace{-2mm}
\begin{equation}
\textbf{q}_{i}^S=\sum_{j=1}^{N} \alpha_{ij} (\textbf{W}_v \textbf{q}_j), i=1,\dots, N,
\vspace{-2mm}
\end{equation}
and $\alpha_{ij}$ are self-attention weights calculated by query and key that exploit the connectivity among landmarks. A residual addition and layer normalization are used as those in normal transformer block. The output is renamed as $\textbf{Q}^S$.

The deformable attention layer takes $\textbf{Q}^S$ as query and the memory feature $\textbf{M}$ as value.
Instead of calculating the relationship between each element of $\textbf{Q}^S$ and $\textbf{M}$, the deformable attention~\cite{deformable2020} only attends to a small set of features, obtained by sampling $\textbf{M}$ according to sampling points. The calculation is formulated as
\vspace{-2mm}
\begin{equation}
\label{qid}
\vspace{-2mm}
\textbf{q}_{i}^D=\sum_{k=1}^{K} \beta_{ik} (\textbf{W} \textbf{x}_{ik}), i=1,\dots, N, 
\end{equation}
where $\textbf{x}_{ik}$ are image features sampled from $\textbf{M}$. $K$ is the entire sampling number. The sampling locations for $\textbf{q}_{i}^D$,
denoted as $\textbf{p}_{ik} \in \mathcal{R}^2$, are calculated by $\textbf{p}_{ik} = \textbf{r}_{i} + \delta
\textbf{p}_{ik}$, where $\textbf{r}_{i}$ denotes the reference point, which is the $i$-th landmark coordinate
calculated from the previous decoder layer, and $\delta \textbf{p}_{ik}$ are sampling offsets, obtained via
linear projection over the query feature $\textbf{q}_{i}^S$. $\beta_{ik}$ denotes the attention weights over the
sampling features, which are calculated by another linear projection over $\textbf{q}_{i}^S$, and a softmax
operation. The sampling process extracts more related landmark features from multi-level feature maps, which reduces
the feature searching area to a large extent and accelerates model convergence. A residual addition, layer normalization
and feed-forward network are followed, and the output is re-denoted as $\textbf{Q}^D = [\textbf{q}_1^D, \dots, \textbf{q}_N^D]$.

The final projection is computed by offset predictor (a 3-layer perceptron in this paper). It takes $\textbf{Q}^D$ as input and predicts the coordinate offsets $\textbf{Y}^o$ with regard to the reference points $\textbf{R}$. The landmark coordinates are then calculated by,
\begin{equation}
\label{yt}
\textbf{Y}_t=\sigma(\textbf{Y}^o_t + \sigma^{-1}(\textbf{R}_t)),
\end{equation}
where $t$ means for the $t$th decoder layer, $t=1,...,T$. $\textbf{Y}^o_t \in \mathcal{R}^{N \times 2}$ are the
predicted coordinate offsets, $\textbf{R}_t \in R^{N \times 2}$ are coordinates of reference points and
$\textbf{R}_t=\textbf{Y}_{t-1}$.

Note that the input query matrix $\textbf{Q}$ is also updated by each decoder layer. $\textbf{Q} =
\textbf{Q}_0$ for the first decoder layer, and $\textbf{Q} = \textbf{Q}^D_{t-1}$ for others. $\textbf{Q}^D_{t-1}$  is the output from previous
deformable attention layer.



\noindent{\bf Parallel Decoder.}
DETR and deformable DETR~\cite{DETR2020,deformable2020} employ several layers of encoder to learn more discriminative image features. DTLD removes the encoder module to save
parameters and computational costs. However, experiments show that the encoder is indeed beneficial to detection performance. Instead of inheriting the serial encoder-decoder architecture, we propose a \textit{parallel decoder}, where the memory feature is updated coherently during the decoding process, along with landmark coordinates refinement.
The simple variation improves landmark detection accuracy furthermore.

\begin{figure}[t]
    \centering
        \includegraphics[width=0.9\linewidth]{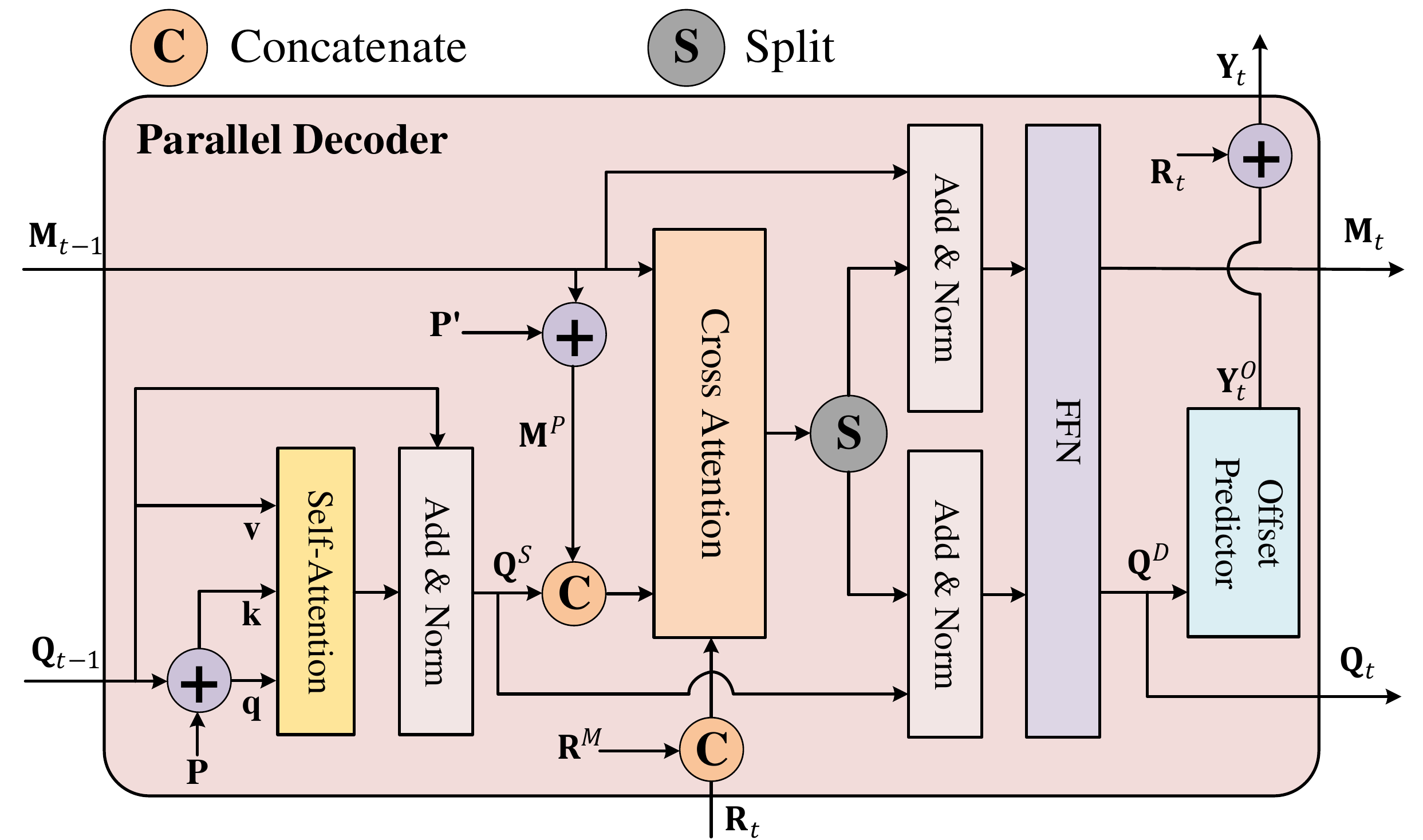}
    \caption{Detailed structure of the proposed parallel decoder. The feature memory \textbf{M} is also updated in the process, sharing the parameters and operations of cross attention and FFN with query \textbf{Q}.}
\label{fig:UTLD_Module}
\vspace{-1mm}
\end{figure}

As shown in Figure~\ref{fig:UTLD_Module}, given the memory feature $\textbf{M}$, we first add both level embedding and pixel position embedding, denoted together as $\textbf{P}'$,
to indicate which level the feature comes from and the spatial location of the feature in feature maps. The embedding added features, denoted as $\textbf{M}^P$, are used as the query for updating image feature, \ie, 
\vspace{-1mm}
\begin{equation}
\label{fj}
\vspace{-1mm}
\textbf{f}_{j}=\sum_{k=1}^{K} \gamma_{jk} (\textbf{W} \textbf{x}_{jk}), j=1,\dots, M. 
\end{equation}
$\textbf{f}_{j}$ are updated image features, $\textbf{x}_{jk}$ are sampled features from $\textbf{M}$ according to sampling location $\textbf{p}_{jk} = \textbf{r}^M_{j} + \delta \textbf{p}_{jk}$. Similarly,
$\delta \textbf{p}_{jk}$ and $\gamma_{jk}$, which denote the sampling offsets and attention weights, are computed by
linear projection over $\textbf{M}^P$. The reference points $\textbf{r}^M_{j} \in [0,1]^2$ are normalized
coordinates of memory feature on each feature map.

In the parallel decoder, we concatenate $\textbf{M}^P$ and $\textbf{Q}^S$ as the overall query features, concatenate
$\textbf{r}^M_{j}, j=1, \dots, M$ and $\textbf{r}_{i}, i=1, \dots, N$ as the reference points, and update both image
features and landmark query features simultaneously according to Eq~\ref{fj} and Eq~\ref{fig:UTLD_Module}.
The layer parameters are shared except that we use separate layer normalizations for image and landmark query. It results in only $1.2K$ more parameters compared to the basic decoder counterpart. The updated image features will be used as the
memory feature next, and the updated landmark query features will be used to calculate the
offsets $\textbf{Y}^o$.

\begin{table*}[ht]
	\newcommand{\tabincell}[2]{\begin{tabular}{@{}#1@{}}#2\end{tabular}}
	\begin{center}
        \setlength{\tabcolsep}{3mm}
		\scalebox{0.85}{
			\begin{tabular}{lcccccccccc}
                \hline
				\multirow{2}*{Method} & \multirow{2}*{Year} &  \multirow{2}*{Backbone} & \multirow{2}*{\tabincell{c}{Pre-  \\ Trained}} & \multicolumn{3}{c}{300W (NME)} & \multirow{2}*{\tabincell{c}{COFW  \\ (NME)}} & \multirow{2}*{\tabincell{c}{AFLW  \\ (NME)}}  & \multicolumn{2}{c}{WFLW-Full}\\
                \cline{5-7} \cline{10-11}
                & & & & Full & Common & Challenge& & &(NME) & ($\text{FR}_{10\%}$) \\
				\hline
				LAB~\cite{LAB} & 2018 & Hourglass & - & $3.49$ & $2.98$ & $5.19$ & $5.58$ & $1.85$ & $5.27$ & $7.56$\\
				Wing~\cite{Wing} & 2018 & ResNet-50 & Y & $-$ & $-$ & $-$ & $5.07$ &  $1.47$ & $4.99$ & $6.00$\\	
				ODN~\cite{ODN} & 2019 & \textbf{ResNet-18} & Y & $4.17$ & $3.56$ & $6.67$ & $5.30$ &  $1.63$ & $-$ & $-$\\
				HGs~\cite{HGs} & 2019 & Hourglass & - & $4.02$ & $3.45$ & $6.38$ & $-$ &  $1.60$  & $-$ & $-$\\
                DeCaFa ~\cite{decafa} & 2019 & Cascaded U-Net & - & $3.39$ & $2.93$ & $5.26$ & $-$ & $-$  & $4.62$ & $4.84$ \\	
				DAG~\cite{DAG2020} & 2020 & HRNet-W18  & Y & $3.04$ & $2.62$ & $4.77$ & $-$ & $-$  & $4.21$ & $3.04$\\
                BarrelNet-18~\cite{BarrelNet} & 2021 & ResNet-18 & Y & $3.22$ & $2.83$ & $4.84$ & $3.24$ & $-$ & $4.42$ & $-$\\
                BarrelNet-101~\cite{BarrelNet} & 2021 & ResNet-101 & Y & $3.09$ & $2.73$ & $4.60$ & $3.10$ & $-$ & $4.20$ & $-$\\
                \hline
                HRNet~\cite{HRNet2019}  & 2019 & HRNet-W18 & Y & $3.32$ & $2.87$ & $5.15$ & $3.45$ & $1.57$  & $4.60$ & $4.64$\\
                AWing~\cite{AWing}& 2019 & Hourglass &N& $3.07$ & $2.72$ & $4.52$ & $-$ & $1.53$  & $4.36$ & $2.84$\\
                AVS ~\cite{AVS} & 2019 & ITN-CPM &N& $3.86$ & $3.21$ & $6.46$ & $-$ & $-$  & $4.39$ & $4.08$\\
                ADA~\cite{ADA} & 2020 & Hourglass &- & $3.50$ & $\textbf{2.41}$ & $5.68$ & $-$ & $-$  & $-$ & $-$\\
                LUVLi~\cite{LUVLi}& 2020 & DU-Net & N & $3.23$ & $2.76$ & $5.16$ & $-$ & $1.39$  & $4.37$ & $3.12$\\
                PIPNet-18~\cite{PIPNet2021} & 2020 & \textbf{ResNet-18} & Y & $3.36$ & $2.91$ & $5.18$ & $3.31$ & $1.48$  & $4.57$ & $-$\\
                PIPNet-101~\cite{PIPNet2021} & 2020 & ResNet-101 & Y & $3.19$ & $2.78$ & $4.89$ & $3.08$ & $1.42$  & $4.31$ & $-$\\
				\hline
                \textbf{DTLD-s} & 2021 & \textbf{ResNet-18} & N & $3.04$ & $2.67$ & $4.56$ & $3.18$ &
                $1.39$  & $4.14$ & $3.44$\\
                \textbf{DTLD} & 2021 & \textbf{ResNet-18} & Y & $\textbf{2.96}$ & $\textit{2.59}$ & $\textit{4.50}$ & $\textit{3.04}$ &
                $\textit{1.38}$  & $\textit{4.08}$ & $\textit{2.76}$\\
                \textbf{DTLD+} & 2021 & \textbf{ResNet-18} & Y & $\textbf{2.96}$ & ${2.60}$ & $\textbf{4.48}$ & $\textbf{3.02}$ &
                $\textbf{1.37}$  & $\textbf{4.05}$ & $\textbf{2.68}$\\
				\hline
			\end{tabular}
		}
	\end{center}
\vspace{-6mm}
\caption{Comparison with SOTA methods on landmark detection accuracy. We report NME ($\%$) on 300W, COFW and AFLW. On WFLW, both NME ($\%$) and FR ($\%$) at the threshold of $10\%$ are reported. Our method achieved the best accuracy on most datasets by simply using ResNet-18 as the backbone, and the second best on 300W-Common subset. DTLD uses the basic decoder, while DTLD+ adopts the parallel decoder. DTLD-s has all parameters trained from scratch. The top methods are coordinate regression-based while the middle ones are heatmap-based. }
\label{Tab:300w}
\end{table*}

\subsection{Training Target}

We simply use $\mathcal{L}_1$ loss between the predicted landmark coordinates and the ground-truth to train the
model, \ie,
\vspace{-2mm}
\begin{equation}
\mathcal{L} = \sum_{t=0}^{T}\left \| \textbf{Y}_t - \hat{\textbf{Y}} \right \|,
\vspace{-2mm}
\end{equation}
where $\textbf{Y}_0$ is computed by Eq~\ref{y0}, and $\textbf{Y}_t, t=1,\dots, T$ are from Eq~\ref{yt}.
$\hat{\textbf{Y}}$ denotes the ground-truth coordinates.

\section{Experiments}
\label{sec:experiments}

In this section, we perform extensive experiments to verify the effectiveness of the proposed method. All the
experiments are conducted on an NVIDIA v100 GPU. The models are implemented by PyTorch.
\subsection{Datasets}

We conduct experiments on a number of popular 2D face landmark detection datasets, including 300W, WFLW, COFW and
AFLW. 
300W~\cite{300W2013} is collected from five facial datasets, consisting of $3148$ training images and $689$ test
images. The test dataset is further divided into 2 subsets, \ie, common set with $554$ images and challenging set
with $135$ images. Each image is annotated with $68$ landmarks.

WFLW~\cite{LAB} is collected from WIDER Face, which includes large variations in pose, expression and occlusion. Each
face is originally annotated by $98$ landmarks, and re-annotated by $68$ landmarks in~\cite{PIPNet2021}. There are $7,500$ images for training and $2,500$ for test. The test set is
further divided into 6 subsets for different scenarios. 

COFW~\cite{COFW2013} contains $1345$ training images and $507$ test images under different occlusion conditions. Each image is annotated by $29$ landmarks, and we also use $68$ landmarks re-annotated by~\cite{COFWReanno} for the cross-domain setting.

AFLW~\cite{AFLW2011} contains $20000$ images for training and $4386$ images for test, providing $19$ landmarks for each face.

CelebA~\cite{celeba2015} is a large-scale attributes
dataset with 202,599 face images in the wild. We only use the images without annotation for training in Section \ref{crossdataeva}.



\subsection{Implementation Details}

For all datasets, the faces are cropped according to the provided bounding boxes firstly, and then resized to $256 \times 256$. In order to retain more context information, the bounding boxes on 300W and WFLW are enlarged by 10$\%$ and 20$\%$, respectively, following previous work~\cite{PIPNet2021}.
Data augmentation is adopted involving translation, horizontal flipping, rotation, occlusion and blurring.
The whole model is trained end-to-end by Adam optimizer for 120 epochs in total. The learning rate is set to 1e-4 initially and then reduced to 1e-5 at $100$th epoch, where the learning rate for backbone is 10 times smaller than the above. By default, we use 3 decoder layers, with a feature dimension of 256 and
8 heads. For each query, we sample 4 features for each head from each level of the feature maps. The configuration will be analyzed in ablation study. We train the model on 1 v100 GPU with a batch size of 16. The reported results are averaged over three runs.

\begin{figure*}[t]
    \centering
        \includegraphics[width=0.6\textwidth]{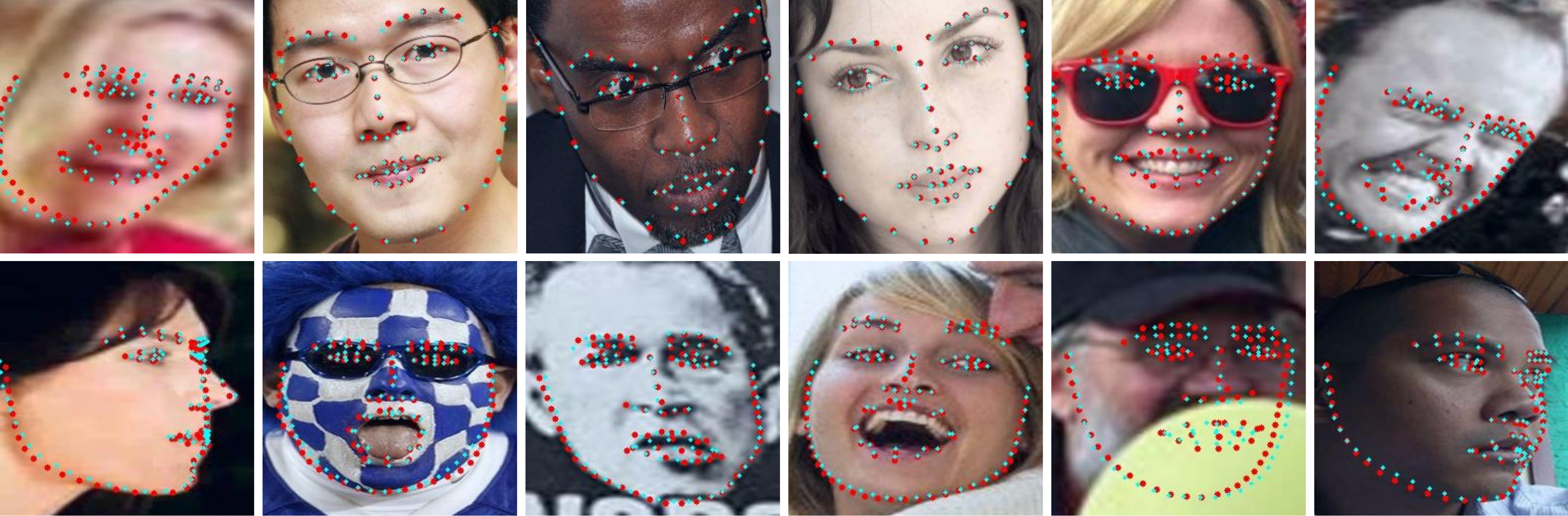}
        \vspace{-2mm}
    \caption{Visualization of typical landmark detection results.
     \textcolor[rgb]{1,0,0}{Red} denotes the ground truth, and \textcolor[rgb]{0,1,1}{cyan} represents our predictions. Our model is able to detect landmarks accurately in various scenarios, such as blur, makeup, expression, occlusion, or even with big pose.}
\label{example}
\end{figure*}

\subsection{Evaluation Metric}

we adopt the most widely used metric, normalized mean error (NME), to evaluate our model for fair comparison with
previous work. It is calculated by,
\vspace{-1mm}
\begin{equation}
\text{NME}(\textbf{Y}, \hat{\textbf{Y}}) = \frac{1}{N}\sum_{i=1}^{N} \frac{\left \| \textbf{y}_i - \hat{\textbf{y}_i}
\right \|_2}{D}.
\vspace{-1mm}
\end{equation}

We employ the prediction from last decoder layer for evaluation. 
$D$ is a normalization distance, and we use inter-ocular
distance for 300W, WFLW, COFW, and image size for AFLW, following common practice.
Failure rate (FR) is also reported which refers to the percentage of failed examples whose NMEs are larger than a certain threshold.

\begin{table*}[htbp]
    \newcommand{\tabincell}[2]{\begin{tabular}{@{}#1@{}}#2\end{tabular}}
    \begin{center}
        \setlength{\tabcolsep}{4mm}
        \scalebox{0.85}{
            \begin{tabular}{lcccccc}
                \hline
                Method & Year &  Backbone & NME($\%$) & Param.(M) & GFLOPs & FPS(GPU) \\
                \hline
                HRNet~\cite{HRNet2019}  & 2019 & HRNet-W18 & $4.60$ & $\textbf{9.7}$ & $4.8$ & $11.7$ \\
                AWing~\cite{AWing}& 2019 & Hourglass & $4.36$ & $25.1$ & $26.7$ & $24.2$ \\
                DeCaFa ~\cite{decafa} & 2019 & Cascaded U-Net & $4.62$ & $\textit{10} $ & $-$ & $32$\\
                LUVLi~\cite{LUVLi}& 2020 & DU-Net  & $4.37$ & $-$ & $-$ & $58.8$ \\
                PIPNet-18~\cite{PIPNet2021} & 2020 & \textbf{ResNet-18} & $4.57$ & $12.0$ & $\textbf{2.4}$ & $\textbf{200}$
                \\
                PIPNet-101~\cite{PIPNet2021} & 2020 & ResNet-101 & $4.31$ & $45.7$ & $10.5$ & $56$ \\
                BarrelNet-18~\cite{BarrelNet} & 2021 & ResNet-18 & $4.42$ & $24$ & $-$ & $\textit{137}$  \\
                BarrelNet-101~\cite{BarrelNet} & 2021 & ResNet-101 & $4.20$ & $56.4$ & $-$ & $55.3$ \\
                \hline
                \textbf{DTLD} & 2021 & \textbf{ResNet-18} & $\textit{4.08}$ & $13.3$ & \textit{2.5} & ${100}$ \\
                \textbf{DTLD+} & 2021 & \textbf{ResNet-18} & $\textbf{4.05}$ & $13.3$ & \textit{2.5} & $78$ \\
                \hline
            \end{tabular}
        }
    \end{center}
\vspace{-5mm}
\caption{Comparison with other methods on Parameter size, GFLOPs and FPS. Our method achieves the highest accuracy with a small amount of GFLOPs and parameters. The FPS is lower than PIPNet-18, which leaves for future improving.}
\label{Tab:speed}
\vspace{-2mm}
\end{table*}


\subsection{Comparison with the SOTA}

As presented in Table~\ref{Tab:300w}, we firstly compare our model with SOTA methods on landmark detection accuracy using the four benchmarks. DTLD is the model with basic decoder, while DTLD-s has all model parameters trained from scratch. DTLD+ is equipped with the parallel decoder.
The models are trained and tested separately, with the default configuration.

The results show that our models consistently outperform all the other methods on all test datasets with a simple backbone. To be specific, our DTLD achieves the NME of $2.96\%$, $3.04\%$ and $1.38\%$ on 300W-Full, COFW and AFLW respectively. In addition, with the NME threshold of $8\%$, the failure rates are $0.29\%$, $0.20\%$ and $0.25\%$ separately.
On WFLW-Full which contains various scenarios, DTLD obtains NME of $4.08\%$, leading to a relative decrease of $2.86\%$ compared to the second best ($4.20\%$ NME), and $7.69\%$ relative to BarrelNet-18 ($4.42\%$ NME), the previous best method using the same backbone. The failure rates are $2.76\%$ at the threshold of $10\%$ and $6.44\%$ at the threshold of $8\%$. Comprehensive results on each WFLW subset are shown in supplementary.

Our model also benefits from the ImageNet pre-trained backbone. Without pre-training (referring to DTLD-s), NMEs increase a lot, but are still smaller than SOTA models with similar model size.
The use of the parallel decoder (DTLD+) improves the detection accuracy further, leading to NME of $4.05\%$ on WFLW and $3.02\%$ on COFW, averagely $0.02\%$ lower than that obtained by DTLD.

Next, we compare the model size and running speed of our models with others. As presented in Table~\ref{Tab:speed}, DTLD has $13.3$M parameters and only $2.5$ GFLOPs, but achieves very competitive accuracy. The running speed is lower than PIPNet-18 and BarrelNet-18 because of the multiple refining process, but is still faster than others. DTLD+ achieves relatively higher accuracy at the sacrifice of running speed.

Some landmark detection results by DTLD are visualized in Figure~\ref{example}. Our model can accurately predict landmarks in the tough scenes for faces with blur, large posture changes, rich expressions, and partial occlusion.

\subsection{Ablation Studies on DTLD}
\label{utld_ex}

We conduct a series of ablation studies to analyze each part of the proposed model. The ablation experiments are performed on WFLW-Full as it includes comprehensive scenarios.

\begin{figure}[t]
    \centering
        \includegraphics[width=0.6\linewidth]{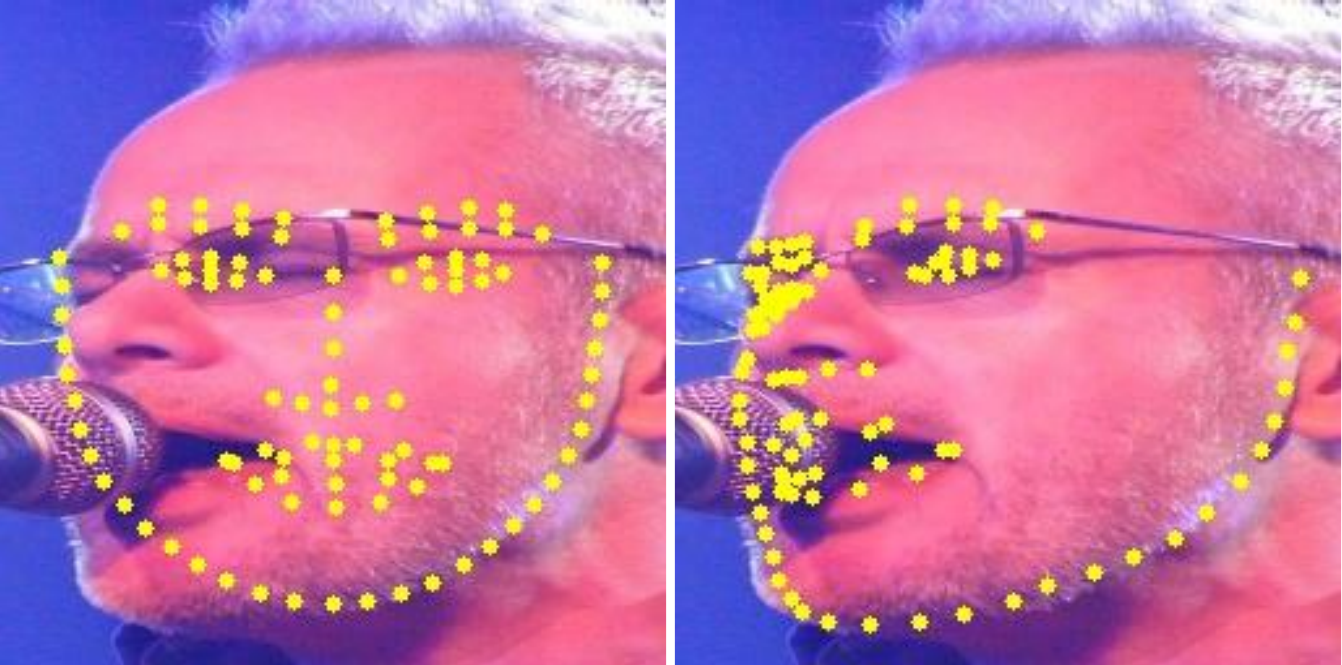}
    \caption{Visualization of the effect of our query initialization. Fiducial landmark positions are produced by randomly initialized $\textbf{Q}_0$ (left), while our query initialization method sets up good starting points (right).}
    \vspace{-5mm}
\label{Qeffect}
\end{figure}

\noindent{\bf Effect of $\textbf{Q}_0$.} In DTLD, we use a well-calculated $\textbf{Q}_0$ as the initial
query features and calculate the initial reference points based on $\textbf{Q}_0$. Here, we perform experiments with
a randomly initialized learnt positional encodings as $\textbf{Q}_0$, as that used in~\cite{DETR2020,deformable2020}. Experimental result in Table~\ref{Tab:Abl1} shows a performance drop of $0.11\%$NME on WFLW. We also visualize the effect in Figure~\ref{Qeffect}. As can be seen, our $\textbf{Q}_0$ will lead to image related initial reference points, instead of a fiducial landmark template.

\begin{figure}[!th]
    \centering
        \includegraphics[width=0.9\linewidth]{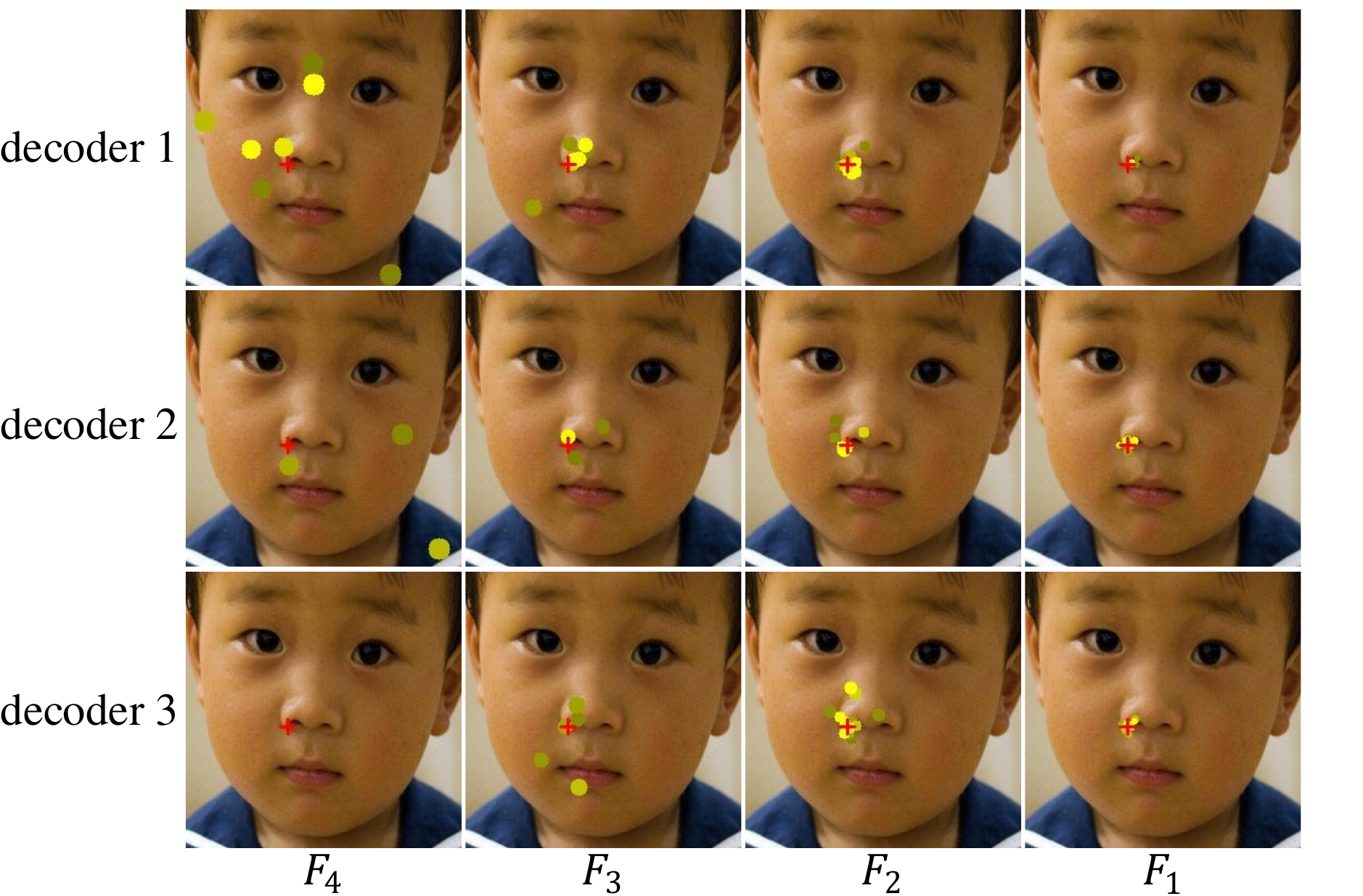}
        \vspace{-2mm}
    \caption{Visualizations of deformable attention on pyramid backbone features. The red cross denotes the ground-truth, while others dots show the sampling points with attention weights expressed by colors. The brighter the point, the greater the weight. We combine the sampling points from all heads for each feature map. Sampling points with attention weights lower than $0.5$ are omitted.}
    \vspace{-1mm}
\label{deformablepoints}
\end{figure}

\begin{table}[ht!]
	\newcommand{\tabincell}[2]{\begin{tabular}{@{}#1@{}}#2\end{tabular}}
	\begin{center}
		\scalebox{0.85}{
			\begin{tabular}{lccc}
				\hline
				Backbone & $\textbf{Q}_0$ & Self-Attn & NME($\%$) \\
				\hline
                R18  & FC &  \CheckmarkBold
                & ${4.08}$ \\
                \hline
                R18 & Random Init. & \CheckmarkBold
                & $4.19$ \\
                R18 & FC & \XSolidBrush
                & $4.17$  \\
                R50 & FC & \CheckmarkBold
                & ${4.07}$  \\
                R101 & FC & \CheckmarkBold
                & ${4.07}$  \\
				\hline
			\end{tabular}
		}
	\end{center}
\vspace{-5mm}
\caption{Ablation study on DTLD, with backbone, $Q$ initial strategy and self-attention analyzed. \emph{R18 / 50 / 101} represent ResNet-18 / 50 / 101 backbone pre-trained by ImageNet . \emph{FC} is our initialization compared with randomly initialized \emph{Random Init}.} 
\vspace{-2mm}
\label{Tab:Abl1}
\end{table}

\noindent{\bf Effect of self-attention.} Self-attention is employed in decoder layer so as to exploit the structural knowledge among landmark positions. Without self-attention, NME of DTLD reduces $0.09\%$ ($4.08\%$ to $4.17\%$) as indicated in Table~\ref{Tab:Abl1}.

\noindent{\bf Effect of backbone.} 
We experiment with different backbones as shown in Table~\ref{Tab:Abl1}. However,
the performance gain is not obvious (only $0.01\%$NME improve) when using deeper backbone like ResNet-50 and ResNet-101, which means DTLD is not sensitive to much deeper backbone.

\noindent{\bf Effect of deformable attention.} We also visualize the multi-scale deformable attention as presented in Figure~\ref{deformablepoints}.
The visualization shows that the deformable module can extract most related image features around the landmark point for coordinate prediction. Moreover, the first decoder layer attends more on the rear feature maps like F3 and F4 that tend to high level global information, while the last decoder layer attends more on the frontal feature maps like F1 and F2 that are apt to capture low level local features for coordinate fine tuning.

\noindent{\bf Effect of model hyper-parameters.} Here we conduct experiments with different model hyper-parameters on DTLD, including the feature dimension $C$ used in decoder
layers, the number of sampling points $K$ used in deformable attention, and the head number used in all attention layers. As shown in Table~\ref{Tab:Abl2}, the higher the feature dimension, the better the performance, but $256$ seems to be enough for feature encoding.  
In our default configuration, for each query feature, we sample $4$ points from each feature level for each head. We then test other numbers such as $2, 6$. However, the change has little impact on the final accuracy, indicating that our model can adaptively decouple the most critical information from redundant features.
We also run models with different head number in both self and deformable attention. More heads benefit final performance.
We visualize the deformable attention for each head in Supplementary. It illustrates intuitively that different heads will pay attention to different directions of image features.

\begin{table}[!t]
	\newcommand{\tabincell}[2]{\begin{tabular}{@{}#1@{}}#2\end{tabular}}
	\begin{center}
		\scalebox{0.85}{
			\begin{tabular}{lccc}
				\hline
				\tabincell{c}{\# Feature  \\ Dimension} & \tabincell{c}{\# Sampling  \\ Points} &  \tabincell{c}{\#
Attention \\ Head}  & NME ($\%$)  \\
                \hline
                256  & 4 & 8 &  ${4.08}$ \\
				\hline
                \textbf{64}  & 4 & 8 &  $4.47$ \\
                \textbf{128}  & 4 & 8 &  $4.29$ \\
                \textbf{512}  & 4 & 8 &  ${4.07}$ \\
				\hline
                256  & \textbf{2} & 8 &  $4.11$ \\
                256  & \textbf{6} & 8 &  $4.09$ \\
                \hline
                256  & 4 & \textbf{4} &  $4.16$ \\
                256  & 4 & \textbf{16} &  ${4.07}$ \\
				\hline
			\end{tabular}
		}
	\end{center}
\vspace{-5mm}
\caption{Ablation on DTLD model hyper-parameters including \emph{feature dimension}, \emph{sampling points}, and \emph{attention heads}. The effect of \emph{sampling points} is tiny, while that of the others is large.}
\label{Tab:Abl2}
\vspace{-2mm}
\end{table}

\begin{table*}[htb!]
	\newcommand{\tabincell}[2]{\begin{tabular}{@{}#1@{}}#2\end{tabular}}
	\begin{center}
		\scalebox{0.85}{
			\begin{tabular}{lccccccc}
				\toprule
                \multirow{7}*{\makecell[c]{DTLD}} &
				\multirow{2}*{\makecell[c]{\# Encoder  \\ Layer}}
				& \multicolumn{6}{c}{\# Decoder Layer}   \\
                \cline{3-8}
                & & 1 & 2 & 3 & 4 & 5 & 6  \\
				\cline{2-8}
				&0 & 4.417 / 12.1 / 165 & 4.187 / 12.7 / 123 & 4.076 / 13.3 / 100 & 4.068 / 14.0 / 82 & 4.044 / 14.6 /
69 & 4.064 / 15.3 / 60 \\
				&1 & 4.369 / 12.6 / 105 & 4.178 / 13.2 / 86 & 4.066 / 13.9 / 71 & 4.026 / 14.5 / 64 & 4.050 / 15.1 /
56 & 4.051 / 15.7 / 51 \\
                &2 & 4.327 / 13.1 / 81 & 4.133 / 13.7 / 69 & 4.047 / 14.4 / 61 & 4.015 / 15.0 / 54 & 4.012 / 15.6 / 45
                & 4.000 / 16.2 / 43 \\	
				&3 & 4.244 / 13.6 / 62 & 4.114 / 14.2 / 54 & 4.028 / 14.8 / 52 & 4.006 / 15.5 / 42 & 3.980 / 16.1 / 35
& 3.978 / 16.7 / 33 \\
				&4 & 4.235 / 14.1 / 53 & 4.079 / 14.7 / 46 & 3.999 / 15.3 / 42 & 3.968 / 16.0 / 34 & 3.972 / 16.6 / 25
& 3.974 / 17.2 / 22 \\
                \hline
				DTLD+ & 0 & 4.417 / 12.1 / 115 & 4.185 / 12.7 / 94 & 4.054 / 13.3 / 78 & 4.022 / 14.0 / 69 & 4.016 / 14.6 /
63 & 3.996 / 15.3 / 55 \\
\bottomrule
			\end{tabular}
		}
	\end{center}
\vspace{-5mm}
\caption{Experimental results on WFLW by using varying encoder and decoder layers. More encoder or decoder layer contributes to higher performance. The last line shows the effect of our proposed parallel decoder. With similar parameters, DTLD+ achieves slightly higher accuracies. The results are demonstrated by NME(\%) / Model Parameter Size (M) /
FPS (on V100 GPU).}
\vspace{-2mm}
\label{Tab:unif}
\end{table*}

\noindent{\bf Effect of decoder.}
Encoder layers are commonly adopted to further encode the image features, \eg, ~\cite{DETR2020,deformable2020,PoseTrans,TFPose}.
Here we also conduct experiments by adding encoder in DTLD as in deformable DETR~\cite{deformable2020} and varying the number of layers in both encoder and decoder. Experimental results in Table~\ref{Tab:unif} show that the added encoder or decoder layers indeed contribute to reduce NME furthermore, even achieving NME smaller than $4\%$ on WFLW. Results also show that the adding of decoder layers has relatively larger effect on accuracy than that of encoder layers. However, the added layer brings more parameters (0.5M for one encoder layer and 0.6M for one decoder layer) and degrades the speed.

To improve the accuracy without increasing model size, we propose the parallel decoder, where the image features are encoded along with the decoding process.
By sharing the deformable attention layers, the model size is almost the same as that without encoder layers (the little parameter increase comes from separate layer normalization), but NME further decreases. When using similar number of parameters, DTLD+ always obtains higher accuracy than DTLD counterparts. When using decoder layers $\geq 3$, DTLD+ gets a higher accuracy compared to DTLD at similar speed.
It should be noted that when we use $1$ parallel decoder layer, the model exactly becomes DTLD with $1$ decoder layer and $0$ encoder. The lower speed is caused by image feature updating which is not used anymore. However, it may provide a chance of inferring occluded face part based on the features so as to improve model performance further. We leave it as a future work.

Moreover, we attempt to remove the backbone and compute the pyramid features simply by image dividing and patch embedding as performed in~\cite{PVT}. The pyramid embeddings are fed into DTLD+ directly for feature encoding and landmark prediction. With 6 layers and only $6$M parameters, our model achieves NME of $4.27\%$ on WFLW.


\begin{table}[!t]
    \centering
    \scalebox{0.85}{
    \begin{tabular}{lccc}
        \toprule
        \multirow{2}*{Methods} & \multicolumn{3}{c}{Test Data}\\
        \cmidrule(){2-4}
        & 300W &COFW68&WFLW68\\
        \midrule
        LAB~\cite{LAB} & 3.49 & 4.62 & $-$ \\
        ODN~\cite{ODN} & 4.17 & 5.30 & $-$ \\
        AVS w/SAN~\cite{AVS} & 3.86 & 4.43 & $-$ \\
        DAG~\cite{DAG2020} & 3.04 & 4.22 & $-$\\
        PIPNet(ST)~\cite{PIPNet2021} & 3.36 & 4.55 & 8.09 \\
        PIPNet(UDA)~\cite{PIPNet2021} & 3.35(-0.3$\%$) & 4.34(-4.6$\%$) & 7.45(-7.9$\%$) \\

        \midrule
        DTLD (ST) & 3.07 & 4.42 & 7.23 \\
        DTLD (UDA) & \textbf{3.03(-1.3$\%$)} & \textbf{4.14(-6.3$\%$)} & \textbf{6.39(-11.6$\%$)} \\
        \bottomrule
        \end{tabular}
    }
    \vspace{-3mm}
    \caption{Cross-dataset evaluation and comparison with others. \emph{ST} means supervised training only on 300W training data, but test on others. \emph{UDA} means unsupervised domain adaption by utilizing COFW and WFLW training images without annotation used.}
    \label{res:crossdataeva}
\end{table}

\begin{table}[!t]
  \centering
  \scalebox{0.85}{
  \begin{tabular}{llccc}
    \toprule
    Methods &Unlabeled Data & 300W & WFLW\\
    \midrule
    \multirow{2}{*}{PIPNet~\cite{PIPNet2021}} &- & 3.36 & $-$  \\
    &CelebA & 3.27 (-2.7$\%$) & $-$ \\
    \midrule
    \multirow{2}{*}{DTLD} &- & 3.07 & 4.08  \\
    &CelebA & \textbf{2.94 (-4.2$\%$)} & \textbf{3.89 (-4.7$\%$)} \\
    \bottomrule
  \end{tabular}
  }
  \vspace{-2mm}
  \caption{Boost our model by using unlabeled images from other domain. Our model shows better scalability, which can be improved more by using unlabeled images. Note that it is the enlarged bounding boxes used in 300W that cause NME of $3.07\%$, larger than $2.96\%$ presented in Table~\ref{Tab:300w}.}
  \label{tab:WFLW_300W}
  \vspace{-1mm}
\end{table}

\subsection{Cross-dataset Evaluation}
\label{crossdataeva}

To verify the robustness and generalization ability of our model, we conduct cross-dataset evaluation on COFW and WFLW testsets, using DTLD trained on 300W training data. To maintain distribution consistency between different datasets, we follow the practice in~\cite{PIPNet2021}, enlarging the provided bounding boxes of 300W, COFW68, and WFLW68 by 30$\%$, 30$\%$ and 20$\%$ respectively. Experimental results in Table~\ref{res:crossdataeva} indicate the robustness of our model in cross dataset evaluation.

In addition, to analyze the model scalability, we perform an unsupervised domain adaption (UDA).  More precisely, we apply the classic self-training strategy and re-train the model using COFW and WFLW training images, without landmark annotation employed. The model trained on 300W is used as a teacher model to reason the pseudo-labels for unlabeled data. They are then combined with the original labeled data and re-train the model. After 3 times of re-training, we achieve NME of $4.14\%$ on COFW68 and $6.39\%$ on WFLW68, new SOTA accuracies on both testsets. Compared to PIPNet, the UDA improvement is more obvious, which demonstrates the good scalability of our method.


Motivated by the good scalability, we attempt to promote the model additionally by leveraging the numerous unlabeled face images from CelebA. With the same self-training paradigm, it is found that the detection accuracy can be further improved on 300W and WFLW-Full testsets.  As indicated in Table~\ref{tab:WFLW_300W}, although the unlabeled images are from a different domain compared with the test datasets, our model can still learn from them and leads to even more accurate landmark prediction.  It finally achieves NME of $2.94\%$ on 300W and $3.89\%$ on WFLW.

Another group of cross-dataset evaluation is performed following the experimental setting in~\cite{LUVLi}. To be specific, we train DTLD from scratch on the training data of 300W Split2, and evaluate it on 300W Split2 test data, Menpo frontal~\cite{Menpo,Menpo1,Menpo2} and COFW68. There are $3837$ images in 300W Split2 train
set and $600$ images in test set.  The $6679$ near-frontal training images in Menpo 2D (denoted as Menpo frontal) are adopted here for evaluation, as well as the $507$ test images in COFW68. Following~\cite{LUVLi}, here we adopt $\text{NME}_\text{box}$ and $\text{AUC}_\text{box}$ as the evaluation metrics. $\text{NME}_\text{box}$ uses the geometric mean of the width and height of the ground-truth bounding box ($\sqrt{ \text{w}_\text{bbox} \cdot \text{h}_\text{bbox} }$) as the normalization distance $\text{D}$. AUC (Area Under Curve) is computed as the area under the cumulative distribution curve, up to a cutoff NME value. The cumulative distribution curve is plotted by the fraction of test images whose NME is less than or equal to the specific NME value on the horizontal axis. Here we use $\text{NME}_\text{box}$ and the cutoff value of $7\%$. The lower the $\text{NME}_\text{box}$, the higher the  $\text{AUC}_\text{box}$, the better the performance. Experimental results are presented in Table~\ref{crossdata2}. Our DTLD without any pretraining achieves the best detection accuracy on 300W Split2 test set and COFW68, even surpassing other methods pretrained on 300W-LP-2D~\cite{300WLP}. On Menpo frontal, our DTLD is still better than previous models without pretraining.

\begin{table}[!t]
    \centering
    \scalebox{0.75}{
    \begin{tabular}{lccc|ccc}
        \toprule
        \multirow{2}*{Methods} & \multicolumn{3}{c|}{$\text{NME}_\text{box}$ (\%) ($\downarrow$) }& \multicolumn{3}{c}{$\text{AUC}^{7}_\text{box}$ (\%) ($\uparrow$)}\\
        \cmidrule(){2-7}
        & 300W & Menpo & COFW68  & 300W &Menpo&COFW68\\
        \midrule
        SAN*~\cite{SAN,SAN2} & 2.86 & 2.95 & 3.50 &59.7&61.9&51.9\\
        2D-FAN*~\cite{FAN} & 2.32 & 2.16 & 2.95 &66.5&69.0&57.5\\
        Softlabel*~\cite{SAN2} & 2.32 & 2.27 & 2.92&66.6&67.4&57.9\\
        KDN~\cite{KDN} & 2.49 & 2.26 & $-$&67.3&68.2& $-$ \\
        KDN*~\cite{KDN} & 2.21 & \textbf{2.01} & 2.73&68.3&71.1&60.1 \\
        LUVLi~\cite{LUVLi} & 2.24 & 2.18 & 2.75&68.3&70.1&60.8 \\
        LUVLi*~\cite{LUVLi} & 2.10 & 2.04 & 2.57&70.2&\textbf{71.9}&63.4 \\

        \midrule
        \textbf{DTLD-s}  & \textbf{2.05} & 2.10 & \textbf{2.47} & \textbf{70.9}& 71.8& \textbf{65.0} \\
        \bottomrule
        \end{tabular}
    }
    \vspace{-3mm}
    \caption{Another group of cross-dataset evaluation. \textbf{DTLD-s} is our proposed DTLD model trained from scratch. The methods marked with * are pretrained on 300W-LP-2D. Our DTLD exceeds previous models without pretraining and is even better than some models pretrained on 300W-LP-2D. }
    \label{crossdata2}
\end{table}

\subsection{Failure Case Analysis}
\label{Sec:limit}

\begin{figure}[t!]
    \centering
        \includegraphics[width=0.95\linewidth]{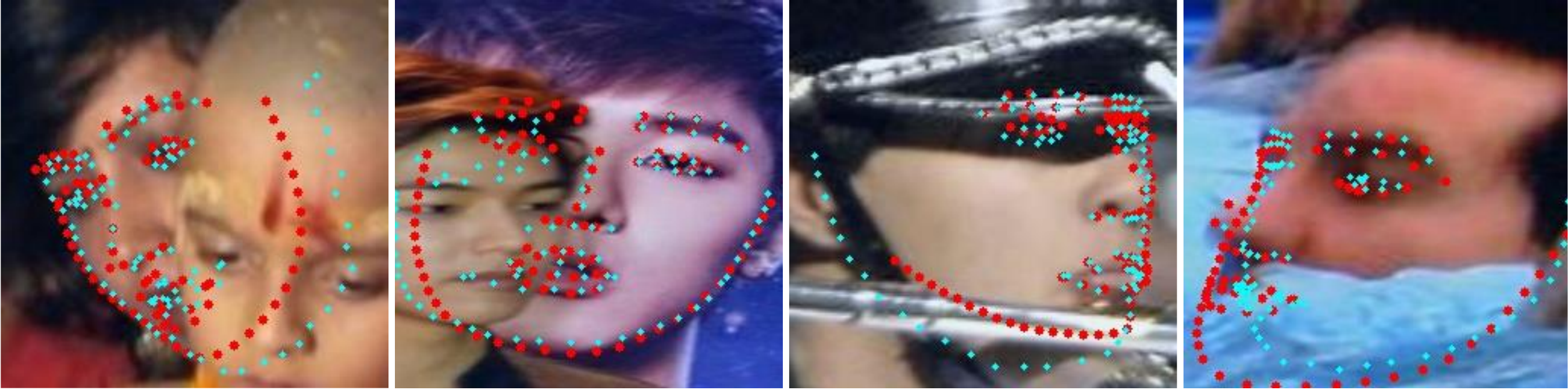}
    \caption{Visualizations of some typical failures. \textcolor[rgb]{1,0,0}{Red} represents the ground truth, and \textcolor[rgb]{0,1,1}{cyan} represents our predictions.}
\label{failure}
\end{figure}

Although our model shows strong superiority on facial landmark detection, it is still weak for face image with severe occlusions, especially obscured by other people, as illustrated in Figure~\ref{failure}. Specifically, 1) If the challenge (\ie, blurring, occlusion, \etc) causes a great uncertainty on face boundary inference, our model may fail. 2) If the face to be aligned is obscured by another face, our model has difficulty in distinguishing the target character, thus leading to large errors. 3) The ambiguity of landmark annotations may lead to poor performance, especially for landmarks on face
boundary.
For these weaknesses, a possible solution is to make better use of the connections between landmarks to infer the invisible part. We leave it as a future work.

\section{Conclusion}
In this paper, we propose an effective and efficient facial landmark detection network DTLD based on cascaded transformer. It directly regresses landmark coordinates and thus can be trained end-to-end.
The use of self-attention and deformable attention in DTLD enables structure relationship exploring and more related image feature extracting. The simple query initialization sets up a better start point for the following refinement.
Moreover, we propose a parallel decoder that refines image features and landmark positions simultaneously, improving the detection performance with few parameter increasing.
Our model achieves new SOTA performance on several standard landmark detection benchmarks, surpassing the other advanced approaches. The running speed is a limitation of current work. Knowledge Distillation based methods may be exploited in the future so as to reduce the cascaded refinement steps and accelerate detection process.


{\small
\bibliographystyle{ieee_fullname}
\bibliography{egbib}
}
\end{document}